\documentclass{bmvc2k}
\usepackage{amsmath}
\usepackage{enumitem}
\usepackage{amssymb}
\newcommand{\hytt}[1]{\texttt{\hyphenchar\font=\defaulthyphenchar #1}}

\title{Adaptive Graphical Model Network for 2D Handpose Estimation}

\addauthor{Deying Kong}{deyingk@uci.edu}{1}
\addauthor{Yifei Chen}{chenyifei.star@gmail.com}{2}
\addauthor{Haoyu Ma}{howiema@seu.edu.cn}{3}
\addauthor{Xiangyi Yan}{11510706@mail.sustech.edu.cn}{4}
\addauthor{Xiaohui Xie}{xhx@ics.uci.edu}{1}

\addinstitution{
 University of California, Irvine}
 
 \addinstitution{Tencent}
 \addinstitution{Southeast University}
 \addinstitution{Southern University of Science and Technology}

\runninghead{D. Kong \bmvaEtAl}{Adaptive Graphical Model Network for 2D Handpose Est.}

\def\eg{\emph{e.g}\bmvaOneDot}

\begin{document}

\maketitle

\begin{abstract}
\vspace{-1em}

In this paper, we propose a new architecture called Adaptive Graphical Model Network (AGMN) to tackle the  task of 2D hand pose estimation from a monocular RGB image. The AGMN consists of two branches of deep convolutional neural networks for calculating unary and pairwise potential functions, followed by a graphical model inference module for integrating unary and pairwise potentials. Unlike existing architectures proposed to combine DCNNs with graphical models,
our AGMN is novel in that the parameters of its graphical model are conditioned on and fully adaptive to individual input images.
Experiments show that our approach outperforms the state-of-the-art method used in 2D hand keypoints estimation by a notable margin on two public datasets. Code can be found at \url{https://github.com/deyingk/agmn}.
\end{abstract}
\section{Introduction}
\label{sec:intro}
Understanding human hand pose is a critical task for many real world AI applications, such as human-computer interaction, augmented reality and virtual reality.
However, hand pose estimation remains very challenging  because the hand is highly articulated and dexterous, and hand pose estimation suffers severely from self-occlusion.
An intuitive approach is to resort to multi-view RGB cameras \cite{simon2017hand, joo2019panoptic}, which unfortunately requires expensive hardware and strict environment configurations.
For practical daily applications, many researchers have also explored the problem under monocular RGB \cite{zimmermann2017learning, panteleris2018using, cai2018weakly} or RGB-Depth  \cite{yuan2018depth, baek2018augmented, wan2018dense} scenarios.
Solving 3D pose estimation problem \cite{zimmermann2017learning, cai2018weakly} often relies on 2D hand pose estimation, making  2D hand pose estimation itself an important task. In this paper, we focus on the task of 2D hand pose estimation from a monocular RGB image.

The advent of Deep Convolutional Neural Networks (DCNNs) has enabled this field to make big progress in recent years. For example, the Convolutional Pose Machine (CPM) \cite{wei2016convolutional} is one of the most successful DCNNs that have been applied to 2D hand pose estimation \cite{simon2017hand}, although it was originally proposed for the task of human pose estimation.
However, despite the fact that DCNNs like CPM have the power to learn good feature representations, they often fail to learn geometric constraints among joints, resulting in joint inconsistency in the final prediction as observed in human pose estimation tasks \cite{song2017thin, ke2018aware}.
For 2D hand pose estimation, the situation could be even worse, since there are more articulations and self-occlusion is severer.

To model the relationships among joints, several studies have also explored the possibility of combining DCNN and the Graphical Model (GM) in pose estimation tasks. Existing methods \cite{tompson2014joint, yang2016end, song2017thin} impose a self-independent GM on top of the score maps regressed by DCNNs. The parameters of the GM are learned during end-to-end training, then these pairwise parameters are fixed and shared by all input images during inference. 

In this paper, we propose the Adaptive Graphical Model Network (AGMN), which is a brand new framework for combining DCNNs and GM. By "adaptive", we mean that the parameters of the GM should be able to adapt to individual input images, instead of being fixed and shared by all input images or among a group of input images. 
The adaptivity of the GM is achieved by setting the pairwise parameters of the GM to be the output of a DCNN whose input is the image.
Meanwhile, the unary potentials (score maps of each hand joint location) are regressed from another branch of DCNN.
Then, final score maps are inferred by the GM using techniques like message passing. The whole AGMN architecture could be trained end-to-end.

We show the efficiency of our proposed framework on two public datasets: the CMU Panoptic Hand Dataset~\cite{simon2017hand} and the Large-scale Multiview 3D Hand Pose Dataset\cite{Francisco2017}. Our approach outperforms the popularly used algorithm CPM by a noticeable margin on both datasets. Qualitative results show our model could alleviate geometric inconsistency among predicted hand keypoints significantly when severe occlusion exists.

The main contributions of this work are:
\begin{itemize}[noitemsep,topsep=0pt,parsep=0pt,partopsep=0pt]
	\item We propose a novel framework integrating DCNNs and graphical model, in which the the graphical mode parameters are fully adaptive to individual input images, instead of being shared among the input images.
	\item By implementing the message passing algorithm as a sequence of 2D convolutions, the inference is performed efficiently and the AGMN could be trained end-to-end.
	\item Our AGMN could reduce the inconsistency and ambiguity of hand keypoints significantly in scenarios of severe occlusion, as shown by experiments on two real world hand pose datasets.
\end{itemize}

%

\section{Related Work}
\label{sec:rel_work}

\subsection{Human pose estimation}
\label{sec:rel_work:pe}
Research on hand pose estimation has benefited from the progress in the study of human pose estimation. On one hand, DCNNs have been successfully applied to human pose estimation \cite{cao2018openpose, he2017mask, fang2017rmpe, zhew2019} in recent years. The DCNN-based algorithms are typically equipped with well crafted deep architectures \cite{simonyan2015vgg, he2016deep} and/or multi-stage training  technique\cite{wei2016convolutional, newell2016stacked}.
Since DCNNs have large receptive fields, they could learn salient and expressive feature representations. However, DCNNs could only capture structural constraints among body parts implicitly, resulting in limited performance in practice when severe occlusion and cluttering exist \cite{song2017thin, ke2018aware}. Some approaches try to learn extra tasks (\eg, offset fields \cite{papandreou2017towards}, compound heatmaps \cite{ke2018aware}) besides heatmaps of joint positions, with the purpose of providing more additional structural information. Nevertheless, these methods still could not fully exploit structural information.


On the other hand,  there is also a trend to combine DCNN with graphical model for pose estimation \cite {tompson2014joint, tompson2014joint, yang2016end}, recently. 
It has been studied in several scenarios, i.e., human pose estimation in a vedio\cite{song2017thin}, multi-person pose estimation \cite{pishchulin2016deepcut, insafutdinov2016deepercut}, multi-person pose tracking \cite{insafutdinov2017arttrack, iqbal2017posetrack} and also in other fields, e.g., word prediction and image tagging\cite{chen2015learning}. 
However, in all of the above approaches, graphical model parameters are not conditioned on individual input images. Authors in~\cite{chen2014articulated} propose to select parameters of graphical models basing on different categories of the relationships between neighboring joints. Although the joint relationships are dependent on input images, all the graphical model parameters are still fixed and shared among different input images after training.  The graphical model parameters are not fully adaptive to individual images. Also the model in~\cite{chen2014articulated} does not support end-to-end training.


\subsection{Hand pose estimation}
The 3D hand pose estimation is a challenging task due to strong articulation and heavy self-occlusion of hands.
Some researcher try to solve the task efficiently with the help of multi-view RGB cameras \cite{simon2017hand, joo2019panoptic}. 
However, this kind of approaches are impractical for daily applications as they require expensive hardware and strict environment configurations.
To circumvent this limitation, other studies have been focused on depth-based solutions \cite{yuan2018depth, baek2018augmented, wan2018dense}  where RGB-D cameras are used.
Due to the ubiquitousness of regular RGB cameras, researchers also have a great interest in solving hand pose estimation from monocular RGB images \cite{zimmermann2017learning, panteleris2018using, cai2018weakly}.

2D hand pose estimation plays an important role in the task of estimating 3D hand pose, since 3D estimation is often inferred from 2D estimation\cite{zimmermann2017learning, panteleris2018using, cai2018weakly}. Current algorithms on 2D hand pose estimation often directly deploy DCNN-based human pose estimators. Among a variety of DCNN-based models, CPM is commonly used in 2D hand pose estimation\cite{simon2017hand, zimmermann2017learning,panteleris2018using, cai2018weakly}, yielding  state-of-art performance. Thus, in this work, we choose CPM as the baseline for comparison with our proposed model.


\section{Method}
\subsection{Basic Framework of Adaptive Graphical Model Network}
\begin{figure}[h!]
\begin{center}
\includegraphics[width=0.75\linewidth]{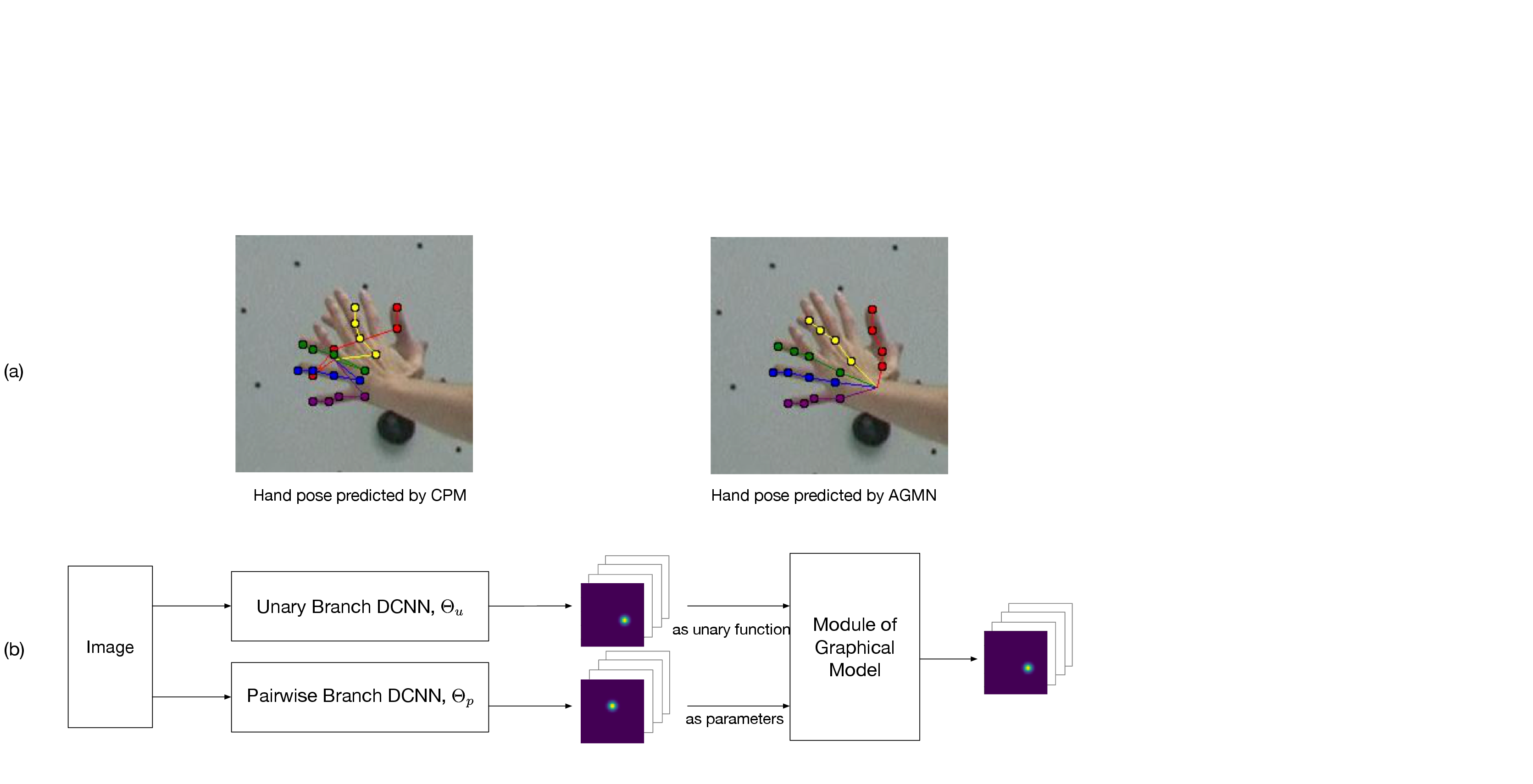}
\end{center}	
  	\caption{Basic flow diagram of adaptive graphical model network.}
\label{fig:model_basic}
\end{figure}

As shown in the left image of Fig.~\ref{fig:model_basic} (a), due to the lack of explicit structural information, CPM fails when the hand is occluded severely, resulting in hand keypoints'  spatial inconsistency. To alleviate this problem, we propose the novel Adaptive Graphical Model Network (AGMN), the efficiency of which could be seen from the right image of Fig.~\ref{fig:model_basic} (a). The model contains two DCNN branches, the  \textit{unary branch} and the  \textit{pairwise branch}, and a graphical model inference module, as depicted in Fig.~\ref{fig:model_basic}~(b). The unary branch would output intermediate score maps of $K$ hand keypoints. Existing DCNN-based pose estimators that regress score maps could be fitted into our AGMN as the {unary branch} easily.
The {pairwise branch} produces parameters that characterize pairwise spatial constraints among hand keypoints.
These parameters would be later used in the graphical model. 
It is the {pairwise branch} that makes our model distinguish from existing models ~\cite{tompson2014joint, chen2014articulated, yang2016end, song2017thin} which also try to combine graphical model with DCNNs. 
In our approach, the parameters of the graphical model are not independent parameters. Instead, they are closely coupled with the input image via a DCNN and they are adaptive to different input images. In approaches from \cite{tompson2014joint, chen2014articulated, yang2016end, song2017thin}, once the graphical model parameters are learned, they would be fixed and used for different input images in future prediction. 

The hand pose estimation problem could be formulated by using a graph. Let $G = (V,  \mathcal{E})$ denote a graph with a vetex set $V$ and an edge set $ \mathcal{E}$, where $V=\{v_1, v_2, \cdots, v_K\}$ corresponds to the set of $K$ hand keypoints and $ \mathcal{E} \subseteq V \times V $ is the set of  edges between neighboring keypoints. Let the discrete variable $x_i\in \mathbb{R}^2$ denote the 2D position of the keypoint associated with $v_i$.

The joint probability distribution of a hand pose configuration is given by


\begin{equation}
p(X | I; \Theta) = \frac{1}{Z} \;  \prod_{i=1}^{|V|} \phi_i(x_i | I; \Theta _{u}) \prod_{(i,j) \in \mathcal{E}} \varphi_{i,j}(x_i, x_j | I; \Theta _{p}),
\label{eq:1}
\end{equation}
where $X = \{x_1, x_2, \cdots, x_K\}$ represents positions of all the keypoints,
$I$ stands for the input image, $|V|$ is the cardinality of the set $V$ and  $Z$ is the partition function.
The whole set of AGMN's prameters $\Theta$ consists of two components, parameters for the {unary branch} and that for the {pairwise branch}, i.e., $\Theta = \{\Theta _{u}, \Theta _{p} \} $. 

\textbf{Unary Terms}. The non-negative term  $\phi_i(x_i | I; \Theta _{u})\in \mathbb{R}$ is the local confidence of the appearance of the $i$-th keypoint at location $x_i$. Let $\mathcal{U}(I; \Theta_u) \in \mathbb{R}^{|V| \times h_u\times w_u}$ denote the output of the unary branch in Fig.~\ref{fig:model}, where  $ w_u$ and $h_u$ are the width and height of the output heatmap. We define 

\begin{equation}
 \phi_i(x_i | I; \Theta _{u}) = \max\left(0, \mathcal{U}^i_{x_i}(I; \Theta _{u})\right),
\label{eq:2}
\end{equation}
where $ \mathcal{U}^i_{x_i}(I; \Theta _{u})\in \mathbb{R}$ is the value of the $i$-th channel of  $\mathcal{U}(I; \Theta _{u})$ evaluated at location $x_i$.

\textbf{Pairwise Terms.} The  term $ \varphi(x_i, x_j | I, \Theta _{p}) \in \mathbb{R}$ represents the pairwise potential function between the $i$-th and $j$-th  keypoints, if $(i, j)$ forms an edge in the graphical model. It encodes spatial constraints between two neighboring keypoints. The pairwise term is given by
\begin{align}
\varphi_{i,j}(x_i, x_j | I; \Theta _{p}) &= \mathcal{F} (x_i, x_j; \theta ^ {(i, j) } ), \\
\theta ^ {(i, j) } &= \max(0, \mathcal{P}^{(i,j)}(I; \Theta _{p})),
\end{align}
where $\mathcal{P}(I; \Theta _{p})\in \mathbb{R}^{|\mathcal{E}|\times h_p \times w_p}  $ is the output of the pairwise branch in Fig.~\ref{fig:model}, 
$ \mathcal{P}^{(i,j)}(I; \Theta _{p}) \in \mathbb{R}^{h_p \times w_p}  $ is a channel of $\mathcal{P}(I; \Theta _{p})$ corresponding to the pair of the $i$-th and $j$-th keypoints. Function $F(\cdot)$ is defined as $\mathcal{F} (x_i, x_j; \theta ^ {(i, j) } ) = \theta ^ {(i, j)} _{x_i - x_j}$, which is an entry of the matrix $\theta ^ {(i, j)}\in\mathbb{R}^{h_p \times w_p} $, indexed by the relative position of the $i$-th keypoint with respect to the $j$-th keypoint.

One can also design $\theta^{(i,j)}$ as a set of parameters of a spring model , and then define $\mathcal{F} (\cdot)$ as a  quadratic function as in~\cite{ chen2014articulated, yang2016end, song2017thin} . In this work we follow the idea in~\cite{tompson2014joint} and design $\theta^{(i,j)}$ to be a 2D matrix, which has a much larger parameter space. 

\textbf{Inference.} The final score maps generated by AGMN are the marginal distributions of $p(X| I; \Theta) $ given in Eq.(\ref{eq:1}). The marginals are defined as
\begin{equation}
p_i(x_i | I; \Theta)=\sum_{V\backslash x_i}{p(X| I; \Theta)},
\end{equation}
which is computed in the {module of graphical model}. Finally, the predicted position of hand keypoint $i$ is obtained by maximizing its marginal probability as 
\begin{equation}
x_i^* = \operatorname*{argmax}_{x_i} p_i(x_i | I; \Theta).
\end{equation}

In summary, the complete parameters in the AGMN model is given by $\Theta = \{\Theta _{u}, \Theta _{p} \} $, consisting the parameters from the unary branch and pairwise branch.

\subsection{Detailed Structure of AGMN}
\label{subsec:detailed}

\begin{figure}[h!]
\begin{center}
\includegraphics[width=0.9\linewidth]{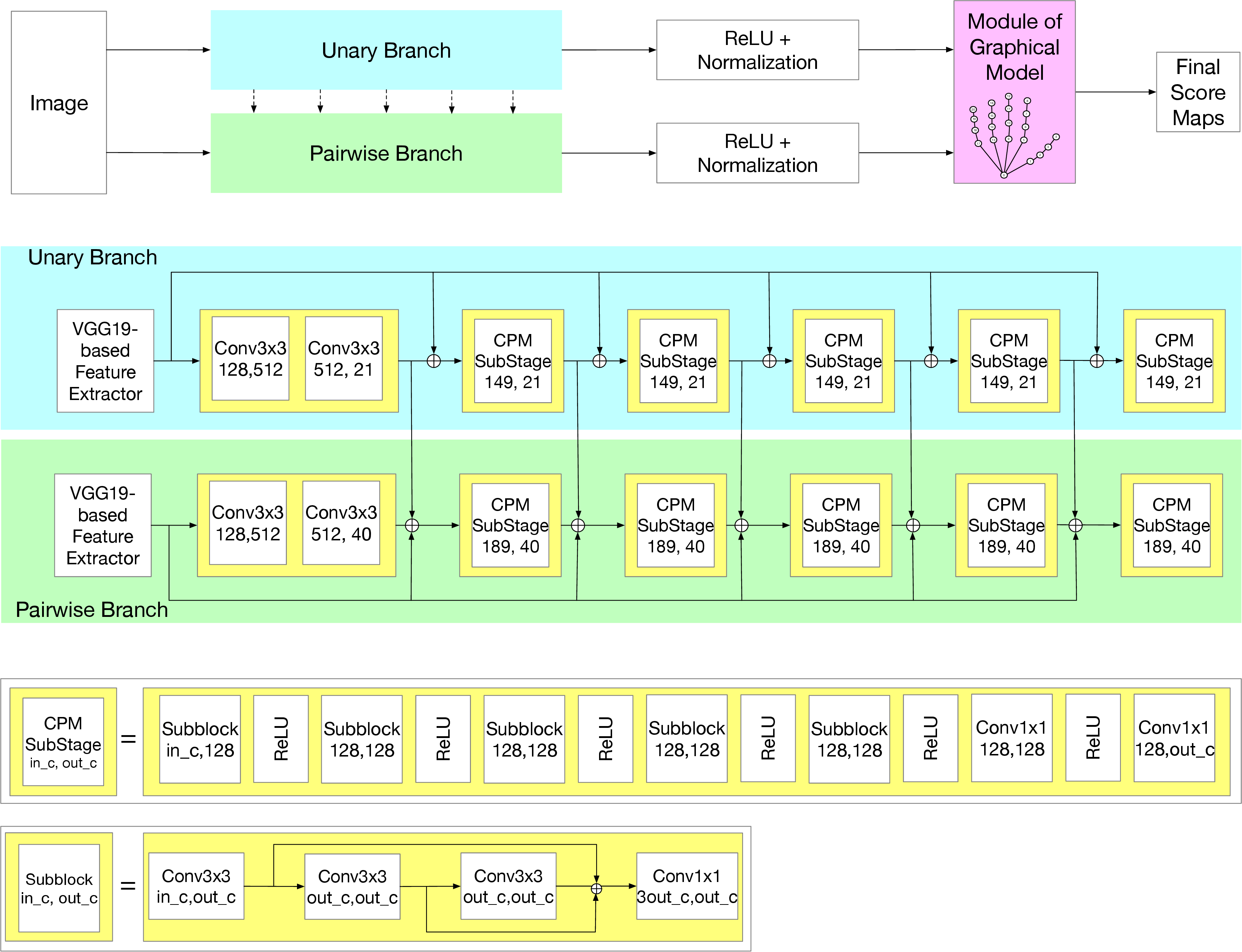}
\end{center}	
  	\caption{More detailed inllustration of adaptive graphical model network.}
\label{fig:model}
\end{figure}

The detailed structure of AGMN is shown in Fig.~\ref{fig:model}.

\textbf{Unary branch.} The CPM architecture in \cite{simon2017hand} is used as the  {unary branch}  in our AMGN model, and it's also compared with as the baseline in our experiments.
A pre-initialized VGG-19 network~\cite{simonyan2015vgg} up to conv4\_4 and additional convolutions are used to produce the 128-channel features, then several prediction stages follows. 
The output of the {unary branch} is a 21-channel score map, each channel corresponding to one keypoint of the hand.

\textbf{Pairwise branch.}  The {pairwise branch} follows the similar structure of the {unary branch}. The only difference is that the {pairwise branch} outputs a 40-channel kernel instead of a 21-channel score map. This 40-channel kernel would later be utilized in the {module of graphical model}. 
There are also some information flowing from the {unary branch} to the {pairwise branch}, as indicated by the arrows between the {unary branch} and {pairwise branch} in Fig.~\ref{fig:model}. We found that adding such information flows would benefit the performance. 

\textbf{Inference.}
\textit{Message Passing.} Sum-product algorithm is widely used for efficient calculation of marginals in a graphical model. Vertices receive messages from and send messages to their neighbors. The sum-product algorithm updates the message sent from hand keypoint $i$ to keypoint $j$ as follows:
\begin{equation}
m_{ij} (x_j) = \sum_{x_i} \; \varphi_{i,j}(x_i, x_j) \phi_i(x_i) \prod_{k \in Nbd(i)\textbackslash j} m_{ki}(x_i).
\label{eq:5}
\end{equation}
Let $M_{ij}$ denote the complete message sent from keypoint $i$ to $j$, then $M_{ij}\in \mathbb{R}^{h_u\times w_u}$,  since $x_j$ could take values from a set of grid points which has the size of $h_u \times w_u$.
After several iterations or convergence, the marginal probabilities are approximated by
\begin{equation}
p_i (x_i) \approx \frac{1}{Z'} \phi_i(x_i) \prod_{k \in Nbd(i)} m_{ki}(x_i),
\label{eq:6}
\end{equation}
where $Z'$ is just a normalization term. 

\begin{figure}[h]
\begin{center}
\includegraphics[width=0.7\linewidth]{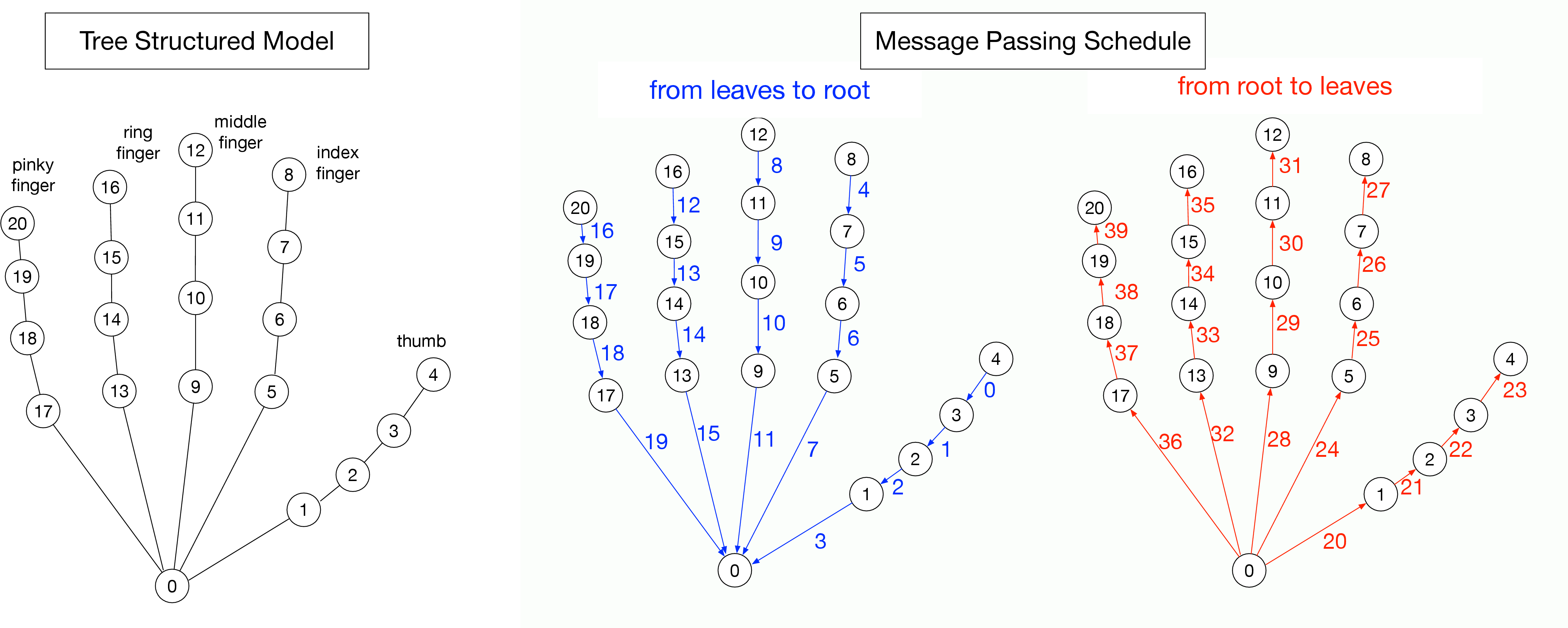}
\end{center}	
  	\caption{Tree structured model and message passing schedule.}
\label{fig:hand_model}
\end{figure}

\textit{Tree Structured Graphical Model}.
In our implemented AGMN, a tree-structured graphical model is used. One advantage of tree-structured model is that exact marginal probability in Eq.(\ref{eq:6}) could be obtained by belief propagation. The tree-structured hand model is shown in Fig.~\ref{fig:hand_model}. By passing messages from leaves to root and then from root to leaves, the exact marginal can be reached. The numbers along side each arrow in Fig.~\ref{fig:hand_model} indicates the schedule of massage updates. In total, we only need pass messages 40 times to obtain exact marginals.

\textit{Message updates as convolution operations.} 
When implementing Eq. (\ref{eq:5}), one way to avoid the for loop in the summation is to use matrix product. However, if we write $\varphi_{i,j}(x_i, x_j)$  compactly into a matrix, the dimension of this matrix is huge. Since $x_i$ and $x_j$ could both take $h^u \times w^u$ different values, The matrix storing $\varphi_{i,j}(x_i, x_j)$ would have the size of $(h^u \times w^u)^2$. To save memories during the inference, we resort to convolution operations when performing message passing.

The message update formula in Eq. (\ref{eq:5}) could be rewritten as 
\begin{align}
m_{ij} (x_j) &= \sum_{x_i} \; \varphi_{i,j}(x_i, x_j) h_i(x_i),  \label{eq:7}\\
h_i(x_i) &= \phi_i(x_i) \prod_{k \in Nbd(i)\textbackslash j} m_{ki}(x_i).
\label{eq:7-8}
\end{align}

We could rewrite Eq. (\ref{eq:7})in a form of 2D convolution, if ${(i, j)\in \mathcal{E}}$,
\begin{equation}
M_{ij}= \theta^{(i,j)} \ast H_i, \;
M_{ji}= \left(\theta^{(i,j)}\right)^T \ast H_j,
\label{eq:11}
\end{equation}
where $M_{ij}\in \mathbb{R}^{h_u \times w_u}$, $\theta^{(i,j)} \in \mathbb{R}^{h_p \times w_p}$, $\theta \in \mathbb{R}^{|\mathcal{E}|\times h_p \times w_p}$, $H_i \in \mathbb{R}^{h_u \times w_u}$. The matrix $H_i$ is the compact matrix formed by values of $h_i(x_i)$.  Appropriate zero padding is required on $H_i$ to make the shape of $M_{ij}$ the same as that of $H_i$. This similar idea is also used in~\cite{tompson2014joint}.
Kernel $\theta^{(i,j)}$ could be interpreted as the probability of where keypoint $j$ would be with respect to keypoint $i$, and it encodes the information of relative positions between the keypoint $i$ and $j$. 
In our implementation in Fig.~\ref{fig:model},  to avoid the transpose operation in Eq.(\ref{eq:11}), we let the pairwise branch produce an output $Q\in \mathbb{R}^{2|\mathcal{E}| \times h_p \times w_p} $ which has $2\times |\mathcal{E}| =40$ channels.

\section{Learning}
Since there are two branches of DCNN in the proposed AGMN, we utilize a 3-stage training strategy. Firstly, the {unary branch} is trained. Then, the {pairwise branch} is trained with the {unary branch} fixed. Finally, the whole AGMN is finetuned end-to-end. 

\textbf{Train unary branch.} The {unary branch} is trained alone first. As in \cite{wei2016convolutional, simon2017hand}, intermediate supervision is used during the training. Each stage of the {unary branch} is trained to repeatedly produce the score maps (or belief maps) for the locations of each of the hand keypoints. The ground truth score map of keypoint $i$, denoted as $S_i^*\in \mathbb{R}^{h_u \times w_u}$, is created by putting a Gaussian peak at its ground truth location. The cost function at each stage $t$ of the {unary  branch} is defined by
\begin{equation}
f_t = \sum_{k=1}^{21}||S_k^{t} - S_k^* ||_F^2,
\label{eq:12new}
\end{equation}
where $S_k^{t} $ is the score map of keypoint $k$ generated by the $t$-th stage in the {unary branch}. Notation $||\cdot||_F$ represents the Frobenius norm which is defined as the square root of the sum of the squares of its elements. If we have $T $ stages, then $S_k^{T} = \mathcal{U}^k(I; \Theta _{u})$ in Eq.(\ref{eq:2}).
By adding up the cost functions at each stage, the final loss function of the {unary branch} is
\begin{equation}
L^{unary}=\sum_{t=1}^{T} f_t.
\end{equation}

\textbf{Train pairwise branch.} The {pairwise branch} is trained with the help of the {unary branch}, since there are some information flowing from the {unary branch} to the {pairwise branch} as shown in Fig.~\ref{fig:model}. Parameters of the {unary branch} are frozen during this training phase. 

The goal of the pairwise branch is to learn relative positions between hand keypoints. The pairwise branch produces an output $Q$ of 40 channels, with each channel corresponding to one directed edge in the message passing schedule. The ideal output (ground truth) $Q^*$ of the {pairwise branch} is computed from relative positions of each pair of neighboring hand keypoints which share a common edge in the tree structure. For example, if the $k$-th directed edge (right side of Fig.~\ref{fig:hand_model}) incidents on two hand keypoints $i$ and $j$, say starting from $i$ to $j$, the relative position of these two keypoints is computed as $r_k = l_j- l_i$, where $l_i$ and $l_j$ are length-2 vectors representing the ground truth positions of the keypoints.
Then, the ground truth of the k-th channel of $Q^*$, i.e., $Q^*_k$, is created by putting a Gaussian peak at the location which is $r_k$ away from the center of the 2D matrix. 

We use a similar loss function as that in training the {unary branch}
\begin{equation}
L^{pairwise} = \sum_{t=1}^{T} \sum_{k=1}^{40}||Q_k^{t} - Q_k^* ||_F^2.
\label{eq:15}
\end{equation}

\textbf{Fine tune the whole AGMN.} Since the final outputs of the AGMN are marginal probabilities, the ground truth for the final score map of keypoint $k$, $\tilde S_k^*$ is set to be the normalized version of $S_k^*$ used in Eq.(\ref{eq:12new}).
The loss function defined by the final score maps is given by
\begin{equation}
L^{last} =  \sum_{k=1}^{21}||\tilde S_k - \tilde S_k^* ||_F^2,
\end{equation}
where $\tilde S_k$ is the $k$-th channel of the  output of the AGMN.

The whole AGMN is fine tuned with a loss function which is a weighted sum of loss functions from the {unary branch}, {pariwise branch} and {module of graphical model} as following
\begin{equation}
L = \alpha_{1} L^{unary} + \alpha_{2} L^{pairwise} + \alpha_{3} L^{last}.
\label{eq:17}
\end{equation}

%
%

\section{Experiments}
In this section, we demonstrate the performance of our proposed algorithm on two real-world hand pose datasets. Comparative analysis is also carried out.
\subsection{Experimental settings}

\textbf{Datasets.} We evaluate our model on two public datasets, the CMU Panoptic Hand Dataset (referred to as "CMU Panoptic")\cite{simon2017hand},  and the Large-scale Multiview 3D Hand Pose Dataset (referred to as "Large-scale 3D") \cite{Francisco2017}. (i) The CMU Panoptic dataset contains 14817 annotations of right hands in images of persons from Panoptic Studio. Since this paper's focus is on hand pose estimation other than hand detection, we cropped image patches of annotated hands off the original images using a square bounding box which is 2.2 times the size of the hand. Then, we randomly split the whole dataset into training set (80\%), validation set (10\%) and test set (10\%). 
(ii) The Large-scale 3D dataset contains 82760 images in total. We follow the same preprocessing procedure on this dataset and take care of the keypoints ordering.  Although this is a 3D dataset, it provides an interface to get 2D annotations by performing projection. The Large-scale 3D dataset is split into training set (60000 images), validation set (10000 images) and test set (12760 images). 

\textbf{Evaluation metric.} We consider the "normalized" Probability of Correct Keypoint (PCK) metric from~\cite{simon2017hand}: the probability that a predicted keypoint is within a distance threshold $\sigma$ of its true location. We use a normalized threshold $\sigma$ which ranges from 0 to 1, with respect to the size of hand bounding box.

\textbf{Implementation  details.} Our experiments are conducted using PyTorch. All images are resized to $368 \times 368$ before fed into the AGMN, yielding a final score map of size $46\times46$ for each keypoint.
Also, after being scaled to [0,1], all the images are then normalized using mean = (0.485, 0.456, 0.406) and std = (0.229, 0.224, 0.225). 

The output shape of the unary branch is designed to be $21 \times 46 \times 46$ as in~\cite{simon2017hand}, corresponding to 21 hand keypoints with heatmap resolution of $46 \times 46$. Meanwhile, we set the resolution of the output from the pairwise branch to be $45 \times 45$, by adding a downsampler after the VGG-19 feature extractor. With the resolution being an odd number, the central entry of the matrix corresponds to the case when the relative position between two hand keypoints is a zero vector.

During training, the batch size is set to 32. 
The \hytt{torch.nn.MSELoss(size\_average =None, reduce=None, reduction='mean')} is utilized for the loss function.  A coefficient of 30 is multiplied to this function to avoid the loss being too small. The gaussian peaks used to generate ground truth during training all have standard deviation of 1. Learning rate is set to 1e-4 when training the {unary branch} and {pairwise branch}. When finetuning the whole AGMN, learning rate is set to 1e-5 and the coefficients in Eq.(\ref{eq:17}) are set to $\alpha_1 = 1, \alpha_2 = 0.1, \alpha_3 = 0.1$. 
\subsection{Results}
\begin{figure}[h]
    \centering
    \subfigure[CMU Panoptic.]{\includegraphics[width=0.48\textwidth]{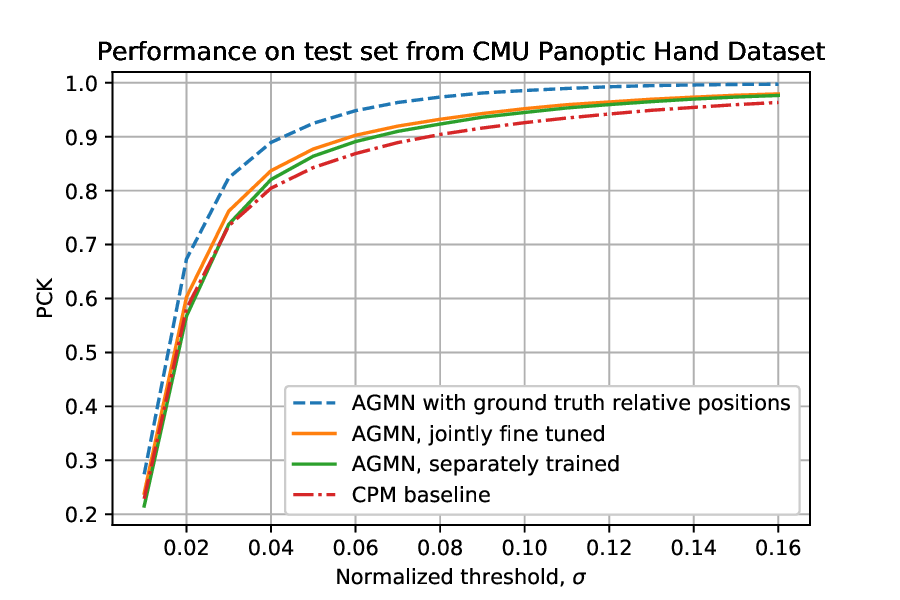}} 
    \subfigure[Large-scale 3D Dataset.]{\includegraphics[width=0.48\textwidth]{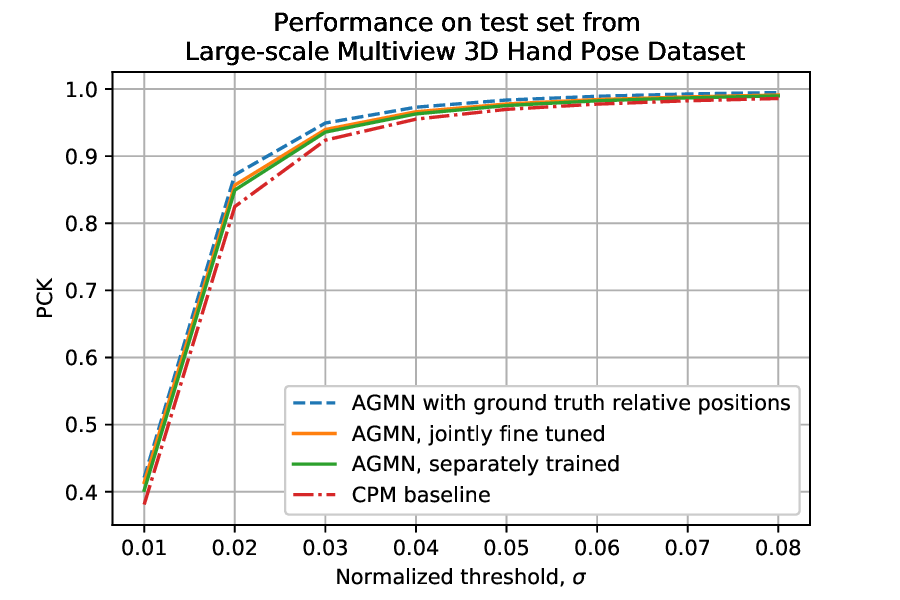}}
    \caption{Model performance.}
    \label{fig:pck_curves}
\end{figure}

Fig.~\ref{fig:pck_curves} shows our model's performance on above mentioned datasets. Detailed numerical results are summarized in Table.~\ref{table:pck}. It is seen that our model outperforms CPM consistently. 

On CMU Panoptic dataset, by training the {unary branch} and {pairwise branch} separately, we see an absolute PCK improvement of 2.12\% at threshold $\sigma=0.05$. A final improvement of 3.45\% is obtained after finetuning the unified AGMN. On Large-scale 3D dataset, our AGMN obtains its highest improvement 3.27\% at thresholds  $\sigma=0.01$. The authors in~\cite{tompson2014joint} stated that "Spatial-Model has little impact on accuracy for low radii threshold". However, based on the results in Fig.~\ref{fig:pck_curves}(b), it is observed that our \textit{adapitve} spatial model has the power of increasing accuracy for low radii threshold. 

The reason why AGMN achieves highest improvement on CMU Panoptic dataset at higher threshold  $\sigma$ than that of Large-scale 3D dataset, probably lies in the fact that CMU Panoptic dataset is a much harder dataset where a lot more occlusions exist.

We also conducted an experiment where the ground truth of the relative positions among hand keypoints ($Q^*$ in Eq.(\ref{eq:15})) are given to the AGMN, with pre-trained {unary branch}. The result of this experiment is actually the upper bound of our AGMN's performance given specific {unary branch}. The result is drawn as the blue dashed lines in Fig.~\ref{fig:pck_curves}. 



\begin{center}
\begin{table}[h]
\resizebox{\columnwidth}{!}{%
\renewcommand{\arraystretch}{1.0}
\begin{tabular}{ c | c c c c c c c c c c c c }
\hline
 Threshold of PCK, $\sigma$& 0.01& 0.02& 0.03& 0.04& 0.05& 0.06& 0.07& 0.08& 0.09& 0.10\\
 \hline
  \multicolumn{11}{c}{CMU Panoptic Hand Dataset}\\
  \hline
 CPM Baseline (\%) & 22.88& 58.10& 73.48& 80.45& 84.27& 86.88& 88.91& 90.42& 91.61& 92.61\\ 
 AGMN Sep. Trained            & 21.52& 56.73& 73.75& 82.06& 86.39& 89.10& 91.00& 92.35& 93.63& 94.50 \\  
 AGMN Fine Tuned   & 23.90& 60.26& 76.21& 83.70& 87.72& 90.27& 91.97& 93.23& 94.30& 95.20\\
 \hline
 Improvement          & 1.02& 2.16& 2.73& \textbf{3.25}& \textbf{3.45}& \textbf{3.39}& \textbf{3.06}& 2.81& 2.69& 2.59\\
 \hline
\hline
 \multicolumn{11}{c}{Large-scale Multiview 3D Hand Pose Dataset}\\
 \hline
  CPM Baseline (\%) & 38.11& 82.48& 92.37& 95.50& 96.97& 97.75& 98.24& 98.58& 98.84& 99.02\\
 AGMN Sep. Trained & 40.22& 84.94& 93.57& 96.29& 97.53& 98.24& 98.68& 98.97& 99.17& 99.34\\ 
 AGMN Fine Tuned   & 41.38& 85.67& 93.96& 96.61& 97.77& 98.42& 98.82& 99.10& 99.29& 99.43\\
 \hline
 Improvement          & \textbf{3.27}& \textbf{3.19}& 1.59& 1.11& 0.80& 0.67& 0.58& 0.52& 0.45& 0.41\\
 \hline
\end{tabular}
}
\caption{Detailed numerical results.}
\label{table:pck}
\end{table}
\end{center}

\begin{figure}[h]
  \centering
\begin{minipage}{1\textwidth}
\includegraphics[width=.15\textwidth]{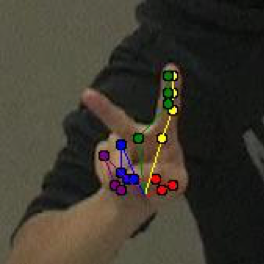}\hfill%
\includegraphics[width=.15\textwidth]{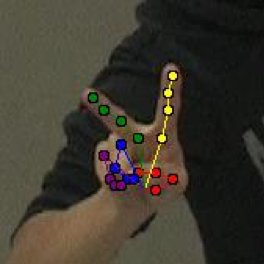}\hfill%
\includegraphics[width=.15\textwidth]{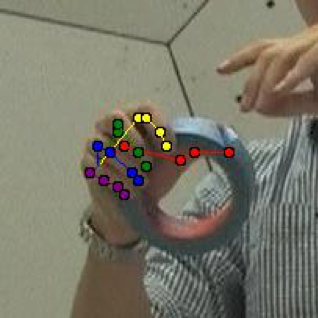}\hfill%
\includegraphics[width=.15\textwidth]{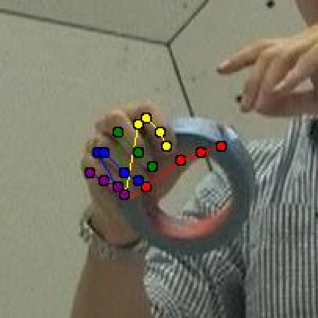}\hfill%
\includegraphics[width=.15\textwidth]{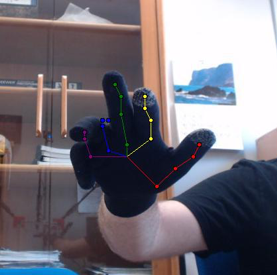}\hfill%
\includegraphics[width=.15\textwidth]{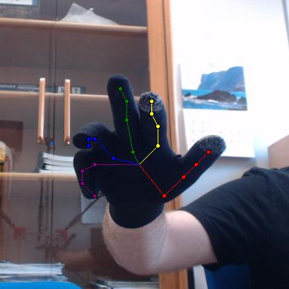}\hfill%
\end{minipage}
\caption{Predicted hand keypoint positions. For each pair of images, the left image shows the result of CPM and the right image shows that of AGMN.}
\label{fig:visualize_result}
\end{figure}


Examples in Fig.~\ref{fig:visualize_result} show that our AGMN could greatly reinforce the keypoints consistency and reduce ambiguities in prediction. Note that the first keypoint in Large-scale 3D dataset is the center of the palm instead of the wrist. 



\section{Conclusion}
This paper provides a new direction on how  deep convolutional neural networks can be combined and integrated with graphical models.
We propose an adaptive framework called AGMN for hand pose estimation, which contains two branches of DCNN, one for regressing the score maps of hand keypoint positions, the other for regressing the parameters of graphical model, followed by a graphical model for inferring the final score maps through message passing. The novelty of our AGMN lies in that the parameters of graphical model are fully adaptive to individual input images, instead of being shared by input images. Experiment results show that the proposed AGMN outperforms the commonly used CPM algorithm on two public hand pose datasets. The proposed framework is general and can also be applied to other deep learning applications where performance can benefit by considering structural constraints. 
\bibliography{egbib}
\end{document}